\documentclass[11pt]{article}

\usepackage[final]{acl}

\usepackage{times}
\usepackage{latexsym}

\usepackage{url}
\usepackage{amsmath,amssymb,amsfonts}
\usepackage{booktabs}
\usepackage{adjustbox}
\usepackage{makecell}
\usepackage{colortbl}
\usepackage{xcolor}
\usepackage{siunitx}
\usepackage[most]{tcolorbox}

\usepackage[T1]{fontenc}

\usepackage[utf8]{inputenc}

\usepackage{microtype}

\usepackage{inconsolata}

\usepackage{bbm}
\usepackage{xspace}
\usepackage{bm}
\usepackage{multirow}

\usepackage{graphicx}

\title{Efficient Process Reward Modeling via Contrastive Mutual Information}

\author{
 \textbf{Nakyung Lee\textsuperscript{1}},
 \textbf{Sangwoo Hong\textsuperscript{2}\thanks{Corresponding authors: Sangwoo Hong; Jungwoo Lee}},
 \textbf{Jungwoo Lee\textsuperscript{1}\footnotemark[1]},
\\
\\
 \textsuperscript{1}Department of Electrical and Computer Engineering, Seoul National University, \\
 \textsuperscript{2}Department of Computer Science and Engineering, Konkuk University,
\\
 \small{
   \textbf{Correspondence:} \href{mailto:email@domain}{swhong06@konkuk.ac.kr, junglee@snu.ac.kr}
 }
}

\begin{document}
\maketitle
\begin{abstract}
Recent research has devoted considerable effort to verifying the intermediate reasoning steps of chain-of-thought (CoT) trajectories using process reward models (PRMs) and other verifier models. However, training a PRM typically requires human annotators to assign reward scores to each reasoning step, which is both costly and time-consuming. Existing automated approaches, such as Monte Carlo (MC) estimation, also demand substantial computational resources due to repeated LLM rollouts. To overcome these limitations, we propose \textbf{contrastive pointwise mutual information (CPMI)}, a novel automatic reward labeling method that leverages the model’s internal probability to infer step-level supervision while significantly reducing the computational burden of annotating dataset. CPMI quantifies how much a reasoning step increases the \textbf{mutual information} between the step and the correct target answer relative to hard-negative alternatives. This \textbf{contrastive} signal serves as a proxy for the step’s contribution to the final solution and yields a reliable reward. The experimental results show that CPMI-based labeling reduces dataset construction time by \textbf{84\%} and token generation by \textbf{98\%} compared to MC estimation, while achieving higher accuracy on process-level evaluations and mathematical reasoning benchmarks. The code is available at \url{https://github.com/nakyungLee20/CPMI}.
\end{abstract}

\section{Introduction}
Recent studies have demonstrated that large language models (LLMs) exhibit remarkable reasoning abilities, particularly when paired with inference-time techniques such as chain-of-thought prompting~\cite{chainofthought} and self-consistency~\cite{selfconsistency}. These advances have motivated growing interest in PRMs, which assess the correctness of intermediate reasoning steps, rather than relying solely on outcome reward models (ORMs) that evaluate final answers. 
However, constructing high-quality process-level supervision data sets for PRM training remains a challenge. Annotating fine-grained step rewards typically requires human experts or high-performing LLMs such as GPT-4, which is both labor-intensive and costly. To address these issues, prior works have explored automated labeling via MC estimation~\cite{mathshepherd, qwenprm}, where multiple reasoning trajectories are sampled and rewards are assigned based on the proportion of trajectories that ultimately yield the correct answer. While this approach eliminates the need for human annotation, it still entails substantial computational overhead, as numerous rollouts are required to obtain low-variance and reliable reward signals. 
This challenge has motivated the development of frameworks that leverage a model's inherent knowledge to reduce such overhead~\cite{OBK}.

In this paper, we propose a novel approach for efficient and automatic step-level rewards labeling by leveraging a model’s internal certainty through likelihood estimates. We hypothesize that modern pretrained LLMs already encode substantial mathematical knowledge. Under this assumption, we posit that quantifying \emph{contrastive} changes in the predicted probability of correct versus incorrect answers at each reasoning step can provide a meaningful reward signal. This formulation eliminates the need for laborious human annotation and the costly rollouts required by MC estimation. Using this reward design, we curate \textbf{CPMI-80k}, a step-level supervision dataset for training verifier models derived from Math-Shepherd~\cite{mathshepherd}. We extract an $80k$ subset of question–solution pairs, automatically annotate scalar rewards for each step along the solution paths.

\begin{figure*}[t]
\begin{center}
\centerline{\includegraphics[width=\textwidth]{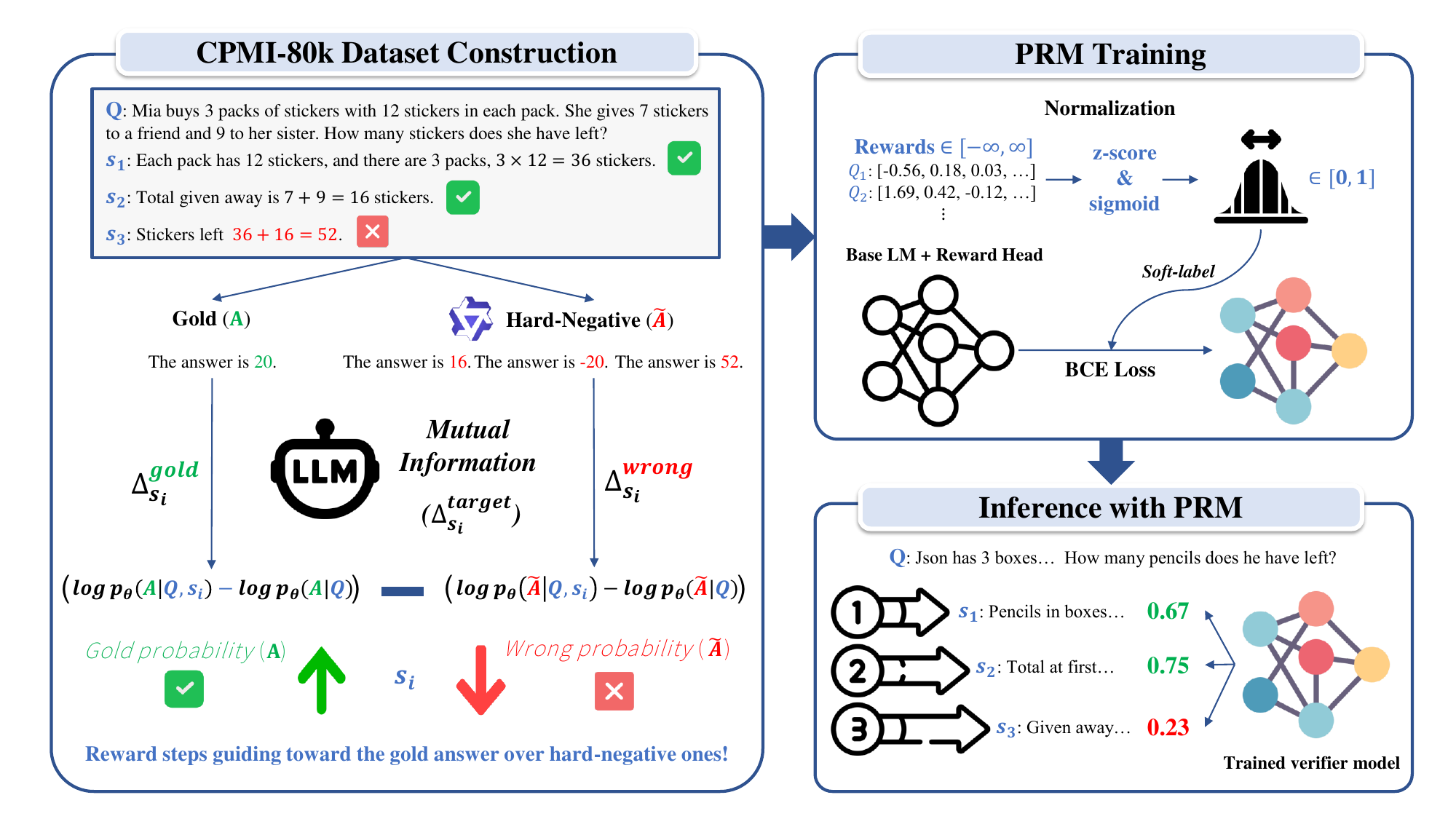}}
\caption{Main framework of our reward modeling and PRM training. We sample both gold and hard-negative answer and compute the logit differences between the with-step and without-step through a simple forward pass for each answer. The CPMI reward for each step is defined as the difference between these logit gaps, quantifying how much the step shifts the model’s belief toward the gold answer while diverging from the wrong answer. After normalizing the CPMI-based rewards and using them as soft labels, we train the PRM with a binary cross-entropy objective. The trained PRM is subsequently used during inference to evaluate or rerank reasoning trajectories.}
\label{fig:main}
\end{center}
\vspace{-0.3in}
\end{figure*}

Empirically, we show that using CPMI-based labeling can assign reliable reward signals that effectively distinguish correct from incorrect reasoning steps. Moreover, it encourages the model to internalize coherent reasoning trajectories during inference, leading to improved performance on mathematical reasoning benchmarks. Compared to MC, our method further reduces the time and computational cost of process-level reward generation by more than 80\%, demonstrating its efficiency and practical scalability. 
\vspace{0.1in}

\textbf{Contributions}
\vspace{-0.1in}
\begin{itemize}
    \item We propose \textbf{contrastive pointwise mutual information (CPMI)}, a new automatic step-level reward labeling approach that removes the need for human annotation and substantially reduces the supervision cost without requiring multiple rollouts.
    \vspace{-0.1in}
    \item We construct \textbf{CPMI-80k}, an automatically labeled step-level reward dataset using our CPMI framework, which provides informative scalar signals for process-level reasoning.
    \vspace{-0.2in}
    \item Through extensive experiments, we show that PRMs trained with CPMI-based labels outperform MC-based approaches on process-level evaluation benchmarks, and further improve mathematical reasoning performance when combined with Best-of-$N$ inference.
\end{itemize}

\section{Related works}
We review prior work along two aspects, (1) reward design for PRMs, and (2) applying trained PRMs to improve mathematical reasoning at inference time.

\paragraph{Reward design for training PRMs} A prevalent strategy for constructing step-level rewards is MC estimation, which samples multiple trajectories and attributes reward by the fraction that ultimately reaches a correct answer~\cite{mathshepherd}. Building on this foundation, several studies have proposed improved labeling or filtering mechanisms. For instance, ~\citet{KHS} proposes evaluation frameworks to assess the accuracy of individual reasoning steps, and ~\citet{rewardingprocess} reformulates step rewards as the difference between MC estimates before and after a step, analogous to the advantage function in reinforcement learning. Meanwhile, ~\citet{epic} and ~\citet{stepwiser} refine the data construction process through efficient annotation and stepwise verification. To further improve label quality, ~\citet{qwenprm} adopts consensus filtering using high-performing LLMs, while \citet{alphamath} introduces a Monte Carlo Tree Search algorithm to annotate step-level rewards without manual supervision. However, all of the above methods still depend heavily on repeated MC rollouts, leading to substantial generation time and computational cost. In contrast, our method requires no MC supervision or uses it only once for initialization and instead relies on the model’s internal log-probabilities, which can be obtained efficiently with a single forward pass.

\paragraph{PRMs for Mathematical Reasoning} Recently, extensive works have focused on improving mathematical reasoning via inference-time strategies such as CoT-style prompting and program-aided computation~\cite{pot, pal, decompose}. When integrating PRMs at inference, one can re-rank or weight candidate traces (e.g., PRM-weighted majority voting or search)~\cite{qwenprm, epic} or inject PRM signals as dense rewards for RL-style training~\cite{rewardingprocess, pure}. Our work falls in the inference-time control regime. We score candidate trajectories with a trained PRM and select the highest average PRM-weighted trajectory as the final solution.

\section{Preliminary}
\paragraph{Notations.} We consider a policy model $\pi_{\theta}$ as a pretrained LLM that generates step-by-step solution trajectories for a given question $q$. Each reasoning step is denoted by $s_i$, representing the $i$-th step in the full solution sequence. The concatenation of all steps up to step $i$ forms a partial solution, denoted as $s_{\leq i}$. We denote the correct answer set as $A$, and an incorrect answer as $\tilde{A}$.

\paragraph{Monte Carlo Estimation.} Given a partial solution $s_{\le i}$, ~\citet{mathshepherd} estimate the step reward using MC rollouts by computing the probability of eventually reaching the correct answer $A$, as defined in Eq.~\eqref{eq:mc}.
\begin{equation}
\label{eq:mc}
r^{(i)}_{\text{MC}} = \frac{1}{T}\sum_{t=1}^{T}
\mathbf{1}\!\left\{ \mathrm{Ans}\!\big (\tau^{(t)} \mid s_{\le i}\big) = A \right\}
\end{equation}
$T$ denotes the number of rollout trajectories. Each continuation trajectory $\tau^{(t)} \sim \pi(\,\cdot \mid s_{\le i})$ is sampled from the policy conditioned on the prefix $s_{\le i}$, $\mathrm{Ans}(\cdot)$ maps a completed trajectory to its final answer, and $\mathbf{1}\{\cdot\}$ is the indicator function.

\paragraph{Process Advantage Verifiers (PAV).} ~\citet{rewardingprocess} introduces an \textit{effective reward} that measures the incremental progress of reasoning between consecutive steps. At each step $i$, PAV combines the MC reward $r^{(i,\theta)}_{\text{MC}}$, obtained from the current policy model $\pi_\theta$, with a \textit{process advantage} term that evaluates the improvement made by a reference policy $\pi_\mu$ relative to the previous step $(i\!-\!1)$.
\begin{equation}
\label{eq:pav}
r^{(i)}_{\text{PAV}}=r^{(i, \theta)}_{\text{MC}} + \alpha \big( r^{(i, \mu)}_{\text{MC}} - r^{(i-1, \mu)}_{\text{MC}} \big)
\end{equation}
This formulation encourages the model to favor reasoning trajectories that exhibit consistent stepwise improvement throughout the intermediate reasoning process.

Using these MC and PAV formulations as Monte Carlo–based baseline methods, we compare them against our proposed reward design.

\section{CPMI: Contrastive Pointwise Mutual Information-based Reward}
\label{sec:intuition}
The approach of quantifying a step’s contribution through \textbf{pointwise mutual information (PMI)} is motivated by a temporal-difference (TD-$\lambda$) perspective~\cite{tdlambda} in reinforcement learning. Unlike standard mutual information which averages over all possible outcomes, PMI utilizes log-probabilities to measure the information gain between specific instances, in our case, how a particular step $s_i$ shifts the likelihood of a specific target $A$ over the prior. This provides a useful analogy for understanding the trade-off between rollout-based and bootstrap-based estimation. Viewed through this TD($\lambda$) lens, the MC reward corresponds to the $\lambda \rightarrow 1$ regime, where long-horizon rollouts serve as full-return targets, whereas the proposed CPMI reward operates in the $\lambda \rightarrow 0$ regime, performing a one-step bootstrap that captures the model’s local belief shift. Thus, to ensure efficiency, we aim to train the verifier relying on bootstrapping (TD-0). 

\subsection{CPMI Reward}
We introduce a \textbf{contrastive} property that explicitly examines: \emph{"How much does a step $s_i$ shift the model's belief toward the gold answer space and away from the negative answer space?"}. Intuitively, an effective step should simultaneously increase the likelihood of producing the correct answer while decreasing likelihood of incorrect ones. By combining temporal bootstrapping and contrastive reasoning perspectives, our formulation enables the reward model to capture fine-grained, discriminative signals that reward genuine reasoning progress and penalize misleading intermediate steps. 

Building on the idea above, we define the CPMI step reward for a given prompt and step $i$, where $M$ denotes the number of incorrect answer alternatives, as shown in Eq.~\eqref{eq:cpmi-marg}. 
\begin{equation}
\label{eq:cpmi-marg}
\begin{split}
    r_{\text{CPMI}}^{i} &= [\underbrace{\log p_{\theta}(A \mid q, s_i) - \log p_{\theta}(A \mid q)}_\textbf{likelihood for gold answer}] - \\
    & \frac{1}{M} \sum_{m=1}^{M} [\underbrace{\log p_{\theta}(\tilde{A} \mid q, s_i) - \log p_{\theta}(\tilde{A} \mid q)}_\textbf{shift toward incorrect answer}]
\end{split}
\end{equation}
The first term quantifies how much the step $s_i$ increases the model’s likelihood of producing the gold answer, while the second term measures its relative shift toward incorrect alternatives. 
To reduce prompt-specific variance, we further average the reward across multiple prompt templates $k \in K$ as shown in Eq.~\eqref{eq:cpmi-final}.
\begin{equation}
\label{eq:cpmi-final}
\hat{r^{i}}_{\text{CPMI}} = \frac{1}{K}\sum_{k=1}^{K} r^{i,k}_{\text{CPMI}}
\end{equation}
This formulation provides an efficient, single-forward-pass estimation of step-level rewards, avoiding the costly MC rollouts.

\subsection{Theoretical Derivation} 
\label{sec:theory}
For each prompt template $k$, we consider the answer distributions $P_k(\cdot) = p_{\theta}(\cdot \mid q_k, s_i)$ and $Q_k(\cdot) = p_{\theta}(\cdot \mid q_k)$, where $q_k$ denotes the instantiated template for the given problem. Our key observation is that the CPMI reward can be interpreted as an approximation to the \emph{Jeffreys divergence} between these two conditional distributions, which can be expressed as
\begin{equation}
\label{eq:jeffrey}
     J(P_k, Q_k) = \mathrm{KL}(P_k\|Q_k) + \mathrm{KL}(Q_k\|P_k).
\end{equation}
Unlike standard KL divergence, this symmetric measure rigorously penalizes discrepancies in both directions. Consequently, it yields a highly sensitive reward signal that intensifies specifically when the distributions occupy distinct regions of the answer space, thereby encouraging the model to clearly distinguish between correct and incorrect reasoning paths. We provide empirical evidence for this behavior in Section~\ref{sec:contrastive}, where we demonstrate that the contrastive signal achieves superior discriminative power compared to non-contrastive counterparts.

In our formulation, positive samples are drawn from $A \sim P_k(A) = p_{\theta}(A \mid q_k, s_i)$ and negative samples from $\tilde{A} \sim Q_k(\tilde{A}) = p_{\theta}(\tilde{A} \mid q_k)$. Since mathematical reasoning tasks in GSM8K and MATH typically feature a unique ground-truth answer, the empirical target distribution $T$ can be modeled as a near-deterministic delta distribution. This property justifies the approximation $T \approx P_k$, so that expectations with respect to $T$ can be estimated using the single sample. Moreover, our controlled sampling setup ensures that the samples from $P_k$ and $Q_k$ satisfy the IID assumption with respect to their underlying model distributions. All samples are generated independently via ancestral decoding, and both $P_k$ and $Q_k$ are computed using the same model and tokenizer under identical temperature and vocabulary settings. This alignment guarantees that the two conditional distributions are normalized over the same support and assign probability mass to a shared answer space, enabling a valid contrastive comparison.

Under this assumption, the CPMI reward in Eq.~\eqref{eq:cpmi-marg} can be rewritten as
\begin{equation}
\label{eq:theory}
\begin{split}
    r^{i,k}_{\text{CPMI}} &\approx \mathbb{E}_{A \sim P_k} \big[ \log \frac{P_k}{Q_k} \big] - \mathbb{E}_{\tilde{A} \sim Q_k} \big[ \log \frac{P_k}{Q_k} \big] \\
    & = \mathrm{KL}(P_k\|Q_k) + \mathrm{KL}(Q_k\|P_k) = J(P_k, Q_k),
\end{split}
\end{equation}
showing that CPMI provides a sample-based approximation to the Jeffreys divergence. 

From a reward perspective, this derivation implies that CPMI explicitly favors steps that induce a large, asymmetric shift between the pre-step and post-step answer distributions, rather than merely increasing the likelihood of the correct answer in isolation. Specifically, the step that shift probability mass toward the correct answer and suppress hard-negative answers yield higher CPMI scores. In contrast, uninformative or redundant steps that result in negligible changes to $P_k$ and $Q_k$ are penalized. 

\subsection{CPMI-80k Dataset Construction}
We now construct the \textbf{CPMI-80k} dataset by automatically labeling step-level rewards on reasoning trajectories from the \textsc{Math-Shepherd} dataset.\footnote{Available at \href{https://huggingface.co/datasets/zhuzilin/Math-Shepherd}{zhuzilin/Math-Shepherd.}} The original corpus contains model-generated reasoning trajectories, where each problem is accompanied by a sequence of step-by-step solutions and a binary correctness label. Since \textsc{Math-Shepherd} lacks explicit gold answers, we align each problem with reference solutions from GSM8K~\cite{gsm8k} and MATH~\cite{math} to recover ground-truth answers. Each generated solution is segmented into reasoning steps and assigned dense scalar rewards using our CPMI-based labeling method, implemented with the Qwen3-8B-Base~\cite{qwen3tech} model. The resulting dataset contains 80k step-annotated trajectories, supporting efficient training and evaluation of PRMs with fine-grained reward supervision. Statistics for \textbf{CPMI-80k} are summarized in Appendix Table~\ref{app-tab:cpmi80k}.

\section{Experimental Settings}
\label{sec:exp_setting}
In this section, we explain the practical implementation details of our CPMI-based reward labeling, focusing on the balance between accuracy and computational efficiency, as well as the training configuration used in subsequent experiments. Further practical details, including reward length normalization and prompt diversification with multiple templates are deferred to Appendix~\ref{app:imple}.

\subsection{Contrastive Sample Generation} 
Constructing high-quality hard negatives is crucial for stable contrastive estimation. We combine model-based sampling with lightweight heuristic perturbations to produce plausible yet incorrect answer candidates that reflect realistic reasoning errors. Specifically, we obtain negative samples $\tilde{A}$ from the model itself, following $\tilde{A} \sim p_{\theta}(\tilde{A} \mid q)$ for theoretical consistency discussed in Section~\ref{sec:theory}. We sample $M=4$ wrong candidates from the model, but if the model keeps generating correct answers on easy problems, we apply heuristic edits such as operator substitution or sign inversion to generate semantically coherent but intentionally incorrect alternatives. Details of the ablation study on different choices of $M$ are provided in Appendix~\ref{app:hardneg}. 

\definecolor{effcol}{HTML}{E8F1FB} 
\begin{table*}[t]
\centering
\small
\setlength{\tabcolsep}{3pt}
\resizebox{\textwidth}{!}{%
\begin{tabular}{lrrrrrrrrrr}
\toprule
& \multicolumn{4}{c}{\textbf{Model Quality}} &
\multicolumn{2}{c}{\textbf{Computation Cost (Ratio)}} &
\multicolumn{4}{c}{\cellcolor{effcol} \textbf{RelEff ($\mathbf{\times}$})} \\
\cmidrule(lr){2-5}\cmidrule(lr){6-7}\cmidrule(lr){8-11}
\textbf{Reward Type} & \textbf{AUC} & \textbf{PB} & \textbf{PRMB} & \textbf{MATH} & \textbf{Time} & \textbf{Token} & \textbf{AUC} & \textbf{PB} & \textbf{PRMB} & \textbf{MATH} \\
\midrule
MC & 0.759 & 27.7 & 38.8 & 45.4 & 1.00 & 1.00 & 1.00 & 1.00 & 1.00 & 1.00 \\
PAV & 0.757 & \underline{36.6} & 49.6 & 47.2 & 1.17 & 2.38 & 0.85 & 1.12 & 1.09 & 0.89 \\
\midrule
CPMI  & \underline{0.765} & 34.6 & \underline{58.8} & \underline{48.2} & 0.16 \textcolor{red}{($\downarrow$84\%)} & 0.02 \textcolor{red}{($\downarrow$98\%)} & \cellcolor{effcol}\textbf{6.34} & \cellcolor{effcol}\textbf{7.85} & \cellcolor{effcol}\textbf{9.53} & \cellcolor{effcol}\textbf{6.68} \\
CPMI\_Merge & \textbf{0.766} & \textbf{36.8} & \textbf{60.7} & \textbf{49.4} & 0.30 \textcolor{red}{($\downarrow$70\%)} & 0.18 \textcolor{red}{($\downarrow$82\%)} & \cellcolor{effcol} \underline{3.38} & \cellcolor{effcol} \underline{4.45} & \cellcolor{effcol} \underline{5.24} & \cellcolor{effcol} \underline{3.64} \\
\bottomrule
\end{tabular}%
}
\caption{Efficiency comparison of CPMI-based reward methods with MC. Computation costs (Time, Generated Tokens) are accumulated over a subset of 10K samples. \textbf{RelEff} quantifies the quality–cost trade-off relative to the baseline, as illustrated in Eq.~\ref{eq:rel-eff}. Best results are highlighted in \textbf{bold}, and second-best results are \underline{underlined}.}
\label{tab:effi-cpmi}
\vspace{-0.1in}
\end{table*}

\subsection{Design Choice for Balancing Efficiency} 
We empirically observe that CPMI rewards at the initial reasoning steps tend to be noisy, as the model is far from the answer space and exhibits high uncertainty in early reasoning stages. To stabilize this effect and enhance the overall accuracy of CPMI labeling, we introduce a hybrid reward formulation that merges CPMI with MC rewards, which we refer to as \textsc{CPMI-Merge}. 
This hybrid design leverages the complementary strengths of both signals, as discussed in Section~\ref{sec:intuition} through the lens of TD-$\lambda$. While MC rewards, though sparse and computationally expensive, capture global information by evaluating full rollouts. On the other hand, CPMI rewards provide dense, bootstrapped feedback that refines local belief updates toward the correct answer. As a result, the hybrid design achieves a more balanced bias–variance trade-off between accuracy and efficiency. 

Implementation details of different merging variants are presented in Appendix~\ref{app:switchat}. Empirically, we report the results of merging MC and CPMI rewards at step index 1, which provides comparable or even improved performance over MC-only labeling while significantly reducing the computational cost of reward construction. This configuration demonstrates that a modest degree of early-stage CPMI integration is sufficient to preserve the global supervision of MC rewards while achieving a favorable balance between performance and efficiency.

\subsection{PRM Training}
We train the PRM using Qwen3-4B-Base~\cite{qwen3tech} as both the policy model during inference and the PRM backbone. A two-layer linear reward head with a non-linear activation is attached to the backbone to output scalar type step rewards. The PRM is optimized using a binary cross-entropy (BCE) loss, as shown in Eq.~\eqref{eq:bce}, where $p_i$ denotes the predicted probability for step $i$, and $\tilde{r_i}$ represents the CPMI-based soft supervision target. 
\begin{equation}
\small
\label{eq:bce}
\mathcal{L}_{\mathrm{BCE}} = -\frac{1}{N}\sum_{i=1}^{N} \Big[ \tilde{r_i} \log p_i + (1 - \tilde{r_i})\log(1 - p_i) \Big]
\end{equation}

Since CPMI scores theoretically range over $\mathbb{R}$ (empirically between $-3$ and $3$), we apply robust z-score normalization to constrain the target values within $[0,1]$ before training as
\begin{equation}
\label{eq:robustz}
\hat{r}_i = \frac{r_i - \mathrm{median}(r)}{\mathrm{MAD}(r) + \epsilon}, \quad \tilde{r_i} = \sigma(\hat{r}_i),
\end{equation}
where $\mathrm{MAD}(r)$ denotes the median absolute deviation, and $\sigma(\cdot)$ is the sigmoid function that maps normalized rewards into the $[0,1]$ scale.

These training configurations closely follow the setup introduced in \textsc{Math-Shepherd}~\cite{mathshepherd} to ensure fair comparison with prior work. This alignment enables consistent evaluation of reward labeling effects across both CPMI-based and MC-based estimation methods.

\paragraph{Baseline Comparison.} For a fair comparison, we directly implement both $r_{\mathrm{MC}}$ (Eq.~\eqref{eq:mc}) and $r_{\mathrm{PAV}}$ (Eq.~\eqref{eq:pav}) using the same backbone model employed for CPMI labeling (Qwen3-8B-Base~\cite{qwen3tech}), and we generate 8 MC rollouts on the same set of 80k question–solution pairs. For $r_{\mathrm{PAV}}$, we additionally load the Qwen2-1.5B-Math model as a prover model, performing 4 rollouts to compute the advantage term, which entails a much higher computational burden than MC methods. 

To further isolate the contribution of CPMI labeling, we introduce a control variant, \textsc{Rand\_Merge}, which preserves the same structure of CPMI\_Merge but replaces CPMI rewards with uniform random values for all steps after the first. This design allows us to disentangle the specific effect of CPMI from other factors in the merging procedure.

\subsection{Evaluation} 
\label{sec:eval}
We assess the efficiency of our proposed reward methods using a \textit{relative efficiency} metric that quantifies the trade-off between model quality and computational cost. Given a quality measure $Q$ and a cost measure $C$, the relative efficiency of a reward method $m$ with respect to a baseline $b$ is defined as
\begin{equation}
\label{eq:rel-eff}
\text{RelEff}(m \mid b) = \frac{Q_m / C_m}{Q_b / C_b},
\end{equation}
where values greater than 1 indicate higher cost-effectiveness. In our experiments, $Q$ corresponds to one of AUC, ProcessBench F1, overall PRMScore of PRMBench or MATH500 accuracy, while $C$ denotes labeling time.

We first evaluate the PRM performance on the \textsc{ProcessBench} benchmark~\cite{processbench} (PB), which measures a model’s ability to detect the first incorrect step within a multi-step reasoning trajectory. To obtain a more fine-grained assessment of step-level abilities, we additionally evaluate on \textsc{PRMBench}~\cite{prmbench} (PRMB), which categorizes reasoning steps into nine dimensions such as consistency and redundancy, enabling a deeper analysis of PRM behavior.
For both benchmarks, thresholds are selected based on the best F1 score on each dataset.

For mathematical reasoning evaluation, we consider two in-domain datasets, GSM8K and MATH500, and two out-of-domain datasets, MMLU-STEM~\cite{mmlu} and Omni-MATH~\cite{omni}. To further examine the PRM’s effectiveness in ranking and selecting high-quality reasoning trajectories, we employ a best-of-N (BoN) test-time scaling strategy. For each problem, the model generates $N$ candidate solution traces, the PRM assigns step scores, and the trajectory with the highest average reward is chosen as the final output, as illustrated in Figure~\ref{fig:main}.

\section{Results}
The main result is reported in Table.~\ref{tab:effi-cpmi}. We first examine the effect of our CPMI reward design, particularly focusing on the role of contrastive targets, in Section~\ref{sec:contrastive}. We then evaluate the overall efficiency and effectiveness of our approach in Sections~\ref{sec:eff}–\ref{sec:bon} with detailed analyses in Appendix~\ref{app-sec:detailed}.

\begin{table}[t]
\centering
\small
\begin{tabular}{lccccc}
\toprule
\textbf{Setting} & \textbf{AUC} & \textbf{PB} & \textbf{PRMB} & \textbf{Math} & \textbf{MMLU} \\
\midrule
\textbf{Gold\_only} & 0.419 & 15.53 & 41.85 & 41.6 & 71.52 \\
\textbf{Neg\_only}  & 0.725 & 24.65 & 57.42 & 40.8 & 68.92 \\
\textbf{Both} & \textbf{0.765} & \textbf{39.48} & \textbf{60.04} & \textbf{50.4} & \textbf{75.67} \\
\bottomrule
\end{tabular}
\caption{Comparison of different reward signal configurations. \textbf{Gold\_only} uses only the correct-answer term, \textbf{Neg\_only} uses only the contrastive negative term, and \textbf{Both} combines both terms as in Eq.~\ref{eq:cpmi-marg}. \textbf{AUC} measures the step-level discriminative ability in a threshold-free manner. \textbf{PB} denotes the average F1 score, \textbf{PRMB} denotes the overall PRMScore, and \textbf{MATH} refers to the MATH500 BoN@8 accuracy.}
\label{tab:gold-neg-m1}
\vspace{-0.15in}
\end{table}

\subsection{Effect of Contrastive Targets}
\label{sec:contrastive}
Prior work on RLVR frequently leverages model-internal likelihood as a confidence signal~\cite{rlp,nover,rlif}. However, Table~\ref{tab:gold-neg-m1} reveals that relying solely on the gold answer (i.e., the first term in Eq.~\ref{eq:cpmi-marg}) introduces a reward-hacking issue. The model may spuriously increase the probability of the correct final answer without learning to reject logically incorrect or misleading steps, which yields poor discrimination on process-level metrics such as ROC-AUC, PB, and PRM. Conversely, using only the negative-target term suppresses the likelihood of wrong answers but still lacks explicit guidance toward the correct reasoning trajectory, resulting in limited performance on downstream tasks such as Math and MMLU.

In contrast, our CPMI objective combines both positive and contrastive negative signals, significantly improving step-level discrimination and leading to better downstream reasoning performance. These results suggest that explicitly penalizing plausible yet incorrect alternatives is essential for mitigating reward hacking and inducing faithful reasoning behavior.

\subsection{Efficiency}
\label{sec:eff}
Table~\ref{tab:effi-cpmi} summarizes the trade-off between model quality and computational cost in constructing process-level datasets. The CPMI method achieves a drastic reduction in construction costs, requiring only $\mathbf{16\%}$ of the total runtime and $\mathbf{2\%}$ of the generated tokens compared to the MC baseline. The efficiency gain is even larger relative to the PAV, as it requires additional rollouts of the prover model. The CPMI\_Merge variant also offers a balanced compromise between performance and budget, maintaining comparable or superior quality to other baselines while reducing both runtime and token usage by more than $\mathbf{70\%}$.

As shown in the last three columns of relative efficiency, both CPMI and CPMI\_Merge consistently achieve over $\mathbf{6\times}$ and $\mathbf{3\times}$ improvements, respectively. These results demonstrate that CPMI-based reward estimation provides a stable and scalable trade-off between cost and performance without substantial quality degradation.

\begin{table}[t]
    \centering
    \small
    \setlength{\tabcolsep}{5pt}
    \begin{tabular}{@{}lcc@{}}
    \toprule
    \textbf{Reward Type} & \textbf{ROC-AUC} ($\uparrow$) & \textbf{Wasserstein} ($\uparrow$) \\ 
    \midrule
    MC  & 0.759 & 0.092 \\
    PAV & 0.757 & 0.105 \\
    \midrule
    CPMI & 0.765 & \textbf{0.135} \\
    CPMI\_Merge & \textbf{0.766} & 0.133 \\
    Rand\_Merge & 0.379 & 0.007 \\
    \bottomrule
    \end{tabular}
    \caption{Distribution-level analysis of PRM scores on the ProcessBench-Math split. Wasserstein reports the 1-Wasserstein distance between the two score distributions in probability space. Bold values indicate the best performance for each metric.}
    \label{tab:dist_analysis}
    \vspace{-0.25in}
\end{table}

\subsection{Process-Level Evaluation}
\begin{figure}[t]
\begin{center}
\centerline{\includegraphics[width=\columnwidth, , trim=4cm 0cm 4cm 0cm, clip]{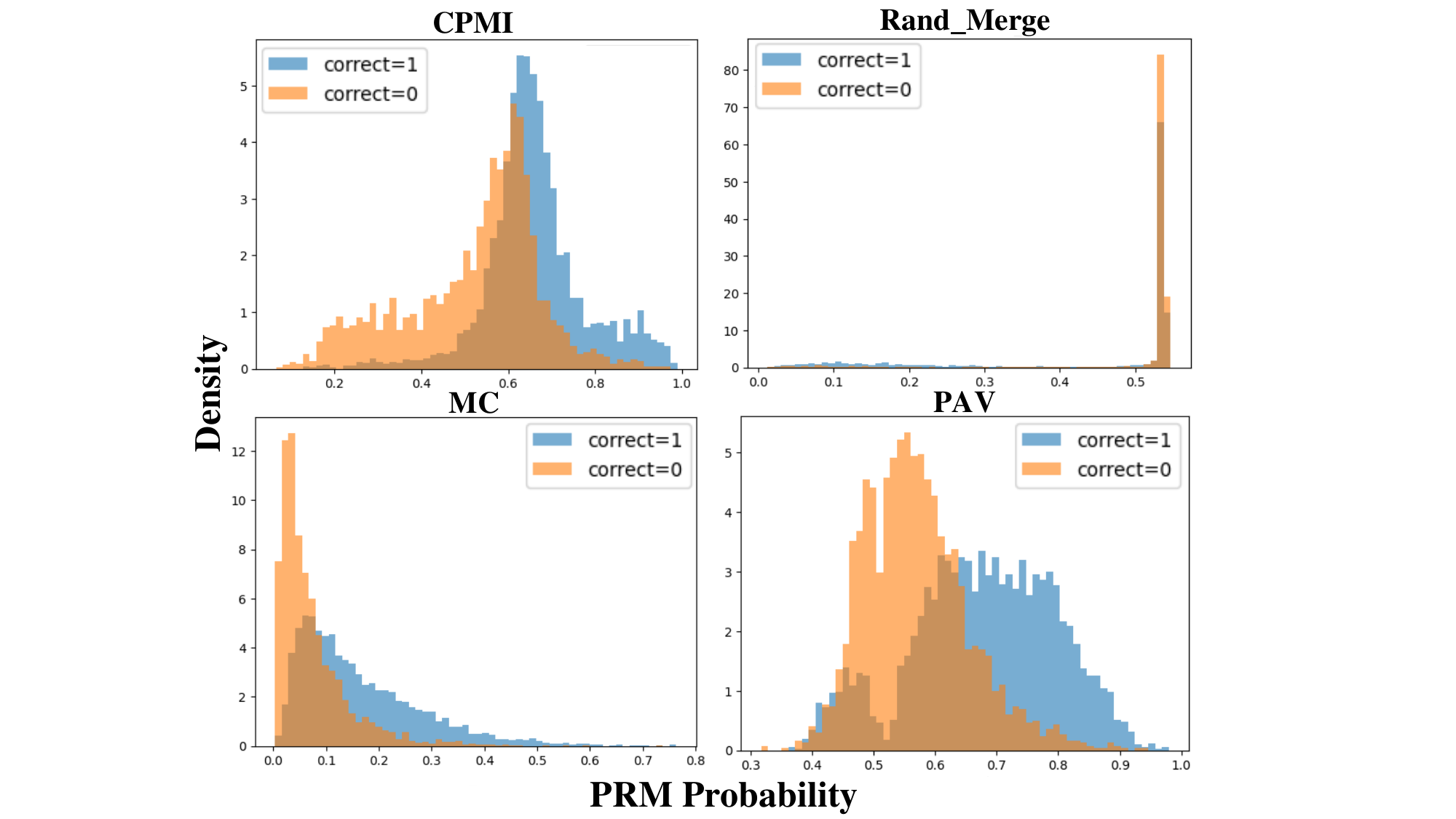}}
\caption{PRM probability distributions on ProcessBench Math split. The x-axis indicates the PRM probabilites. The y-axis shows the normalized density of step occurrences. Blue and orange histograms correspond to correct and incorrect reasoning steps, respectively.}
\label{fig:logitdist}
\end{center}
\vspace{-0.3in}
\end{figure}

\paragraph{ProcessBench.}
\label{sec:pb_analysis}
We evaluate the verifiability of the trained PRM on the ProcessBench. To assign labels, we follow the ProcessBench gold first-error index, treating all steps before the first detected error as correct and all subsequent steps as incorrect. Since this assumption may not perfectly reflect real-world reasoning dynamics, it inevitably produces partially overlapping distributions, as visualized in Figure~\ref{fig:logitdist}. Despite this overlap, the distributional metrics in Table~\ref{tab:dist_analysis} show that CPMI-based rewards achieve higher ROC-AUC compared to MC and PAV, indicating a stronger ability to rank correct steps above incorrect ones. The Wasserstein distance in probability space highlights a clearer difference at the global distribution level. MC exhibits almost no separation between correct and incorrect scores, while CPMI and CPMI\_Merge substantially enlarge this distance by about \textbf{80\%} and exceed PAV. These results indicate that CPMI-based rewards not only preserve step-level discriminative power but also push correct and incorrect steps into more distinct regions of the score space, aligning with our goal of emphasizing steps that move the model toward the gold answer space.

\begin{table}[t]
\centering
\small
\setlength{\tabcolsep}{2pt}
\begin{tabular}{lcccc >{\columncolor{effcol}}c}
\toprule
\textbf{Reward Type} & \textbf{GSM8K} & \textbf{Math} & \textbf{Olympiad} & \textbf{Omni} & \textbf{Average} \\
\midrule
MC          & 35.2 & 38.8 & 17.9 & 18.9 & 27.7 \\
PAV         & 51.8 & \textbf{43.8} & 27.6 & 23.1 & 36.6 \\
\midrule
CPMI        & 49.4 & 39.6 & 24.5 & 24.8 & 34.6 \\
CPMI\_Merge & \textbf{52.0} & 40.8 & \textbf{28.7} & \textbf{25.6} & \textbf{36.8} \\
\midrule
Rand\_Merge & 21.7 & 24.9 & 12.2 & 15.9 & 18.7 \\
\midrule
MS*         & 47.9 & 33.9 & 24.8 & 19.8 & 31.5 \\
RLHFlow*    & 50.4 & 33.4 & 13.8 & 15.8 & 28.4 \\
\bottomrule
\end{tabular}
\caption{Performance by reward type on ProcessBench (F1 only). The \textbf{Average} column indicates the mean F1 across the four datasets. Rows marked with * correspond to
reported performance in \citet{mathshepherd}. Bold values indicate the best F1 in each column.}
\label{tab:pb}
\vspace{-0.2in}
\end{table}

Results in Table~\ref{tab:pb} show that the CPMI-based method consistently outperforms the MC reward across both in-domain and out-of-domain datasets, demonstrating strong generalization in step-level error detection. Moreover, CPMI\_Merge surpasses the PAV baseline while maintaining lower computational cost, confirming the effectiveness of integrating local CPMI signals with global Monte Carlo supervision. Additionally, the large performance gap between CPMI\_Merge and \textsc{Rand\_Merge} highlights that the observed improvements are not merely due to the MC component, but arise from the meaningful CPMI-based supervision. These findings collectively demonstrate that CPMI provides a more stable and informative reward signal for process-level evaluation.

\paragraph{PRMBench.}
\label{sec:prmbench}
We further evaluate step quality across the nine dimensions in PRMBench, grouped into three categories~\cite{prmbench}, as detailed in Section~\ref{sec:eval}. Table~\ref{tab:prmbench} shows that CPMI\_Merge achieves the strongest performance across all category-level averages as well as the overall score. CPMI\_Merge improves the overall PRMBench score by +11.03\% over PAV and +21.87\% over MC, with the largest gains on Soundness, indicating more logically consistent and less redundant step-level judgments. 

\subsection{BoN Inference}
\label{sec:bon}
\begin{figure}[t]
\begin{center}
\centerline{\includegraphics[width=\columnwidth, trim=3.5cm 0cm 3.5cm 0cm, clip]{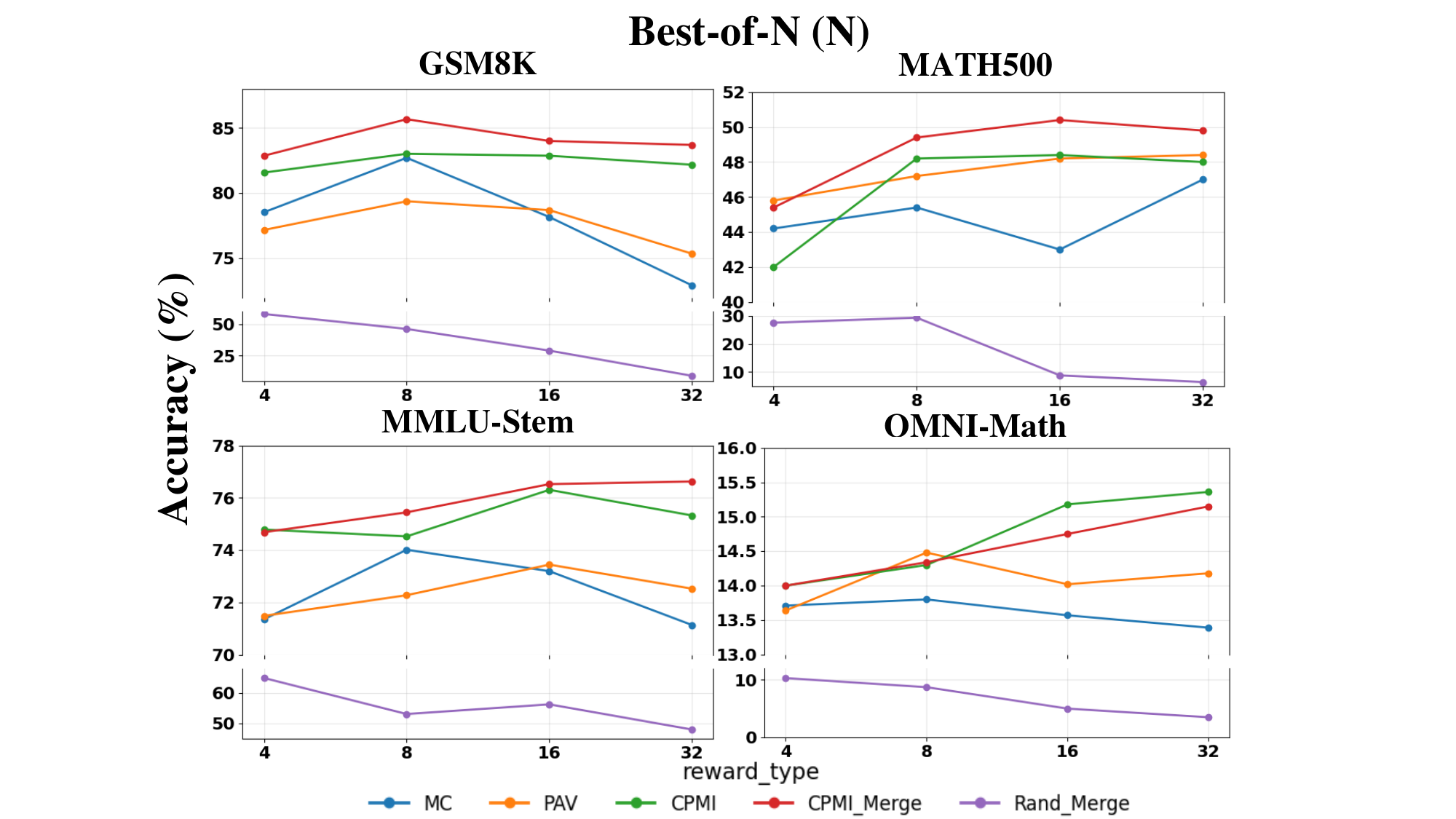}}
\caption{Best-of-$N$ accuracy on math benchmarks. The x-axis denotes the number of samples $N$ used for Best-of-$N$ selection, and the y-axis reports accuracy.}
\label{fig:bon_math}
\end{center}
\vspace{-0.35in}
\end{figure}

Across different values of $N$ and various mathematical reasoning benchmarks, the CPMI-based reward methods, shown in green and red in Figure~\ref{fig:bon_math}, consistently outperform both MC and PAV. As $N$ increases, CPMI and CPMI\_Merge exhibit stable and consistent scaling behavior, suggesting that their PRMs provide a reliable ranking signal for selecting high-quality trajectories under larger sampling budgets. In contrast, MC shows more non-monotonic trends on some datasets (e.g., GSM8K), indicating that noisier step-level supervision can limit the gains from increasing $N$. \textsc{Rand\_Merge} further highlights this effect, as its performance degrades with larger $N$, suggesting that uninformative or misaligned reward signals can be amplified rather than corrected by increased sampling. Overall, these results indicate that BoN inference is highly sensitive to the quality of the underlying reward model. 

Another notable finding from out-of-domain evaluations is that CPMI achieves 76.31\% on MMLU-Stem and 15.18\% accuracy on Omni, compared with 73.45\% and 13.57\% obtained by MC, demonstrating stronger generalization under BoN inference.

\begin{table}[t]
\centering
\small
\setlength{\tabcolsep}{2.5pt}
\begin{tabular}{lccc>{\columncolor{effcol}}c}
\toprule
\textbf{Reward Type} & \textbf{Simplicity} & \textbf{Soundness} & \textbf{Sensitivity} & \textbf{Total} \\
\midrule
MC           & 40.88 & 36.92 & 35.06 & 38.80 \\
PAV          & 47.16 & 49.56 & 47.20 & 49.64 \\
\midrule
CPMI         & 52.56 & 60.21 & 55.59 & 58.75 \\
CPMI\_Merge  & \textbf{54.65} & \textbf{62.30} & \textbf{56.50} & \textbf{60.67} \\
\midrule
Rand\_Merge& 27.65 & 21.64 & 18.65 & 23.20 \\
\bottomrule
\end{tabular}
\caption{PRMBench results. Average scores over nine categories into three step-level dimensions (Simplicity, Soundness, Sensitivity) and the overall average for each reward method.}
\label{tab:prmbench}
\vspace{-0.1in}
\end{table}

\section{Conclusion}
We introduced CPMI, a novel method for automatically generating process-level supervision without relying on human annotation or incurring the substantial computational cost required by traditional MC methods. By providing dense, localized contrastive reward signals, CPMI enables the construction of effective PRMs while reducing generation time by more than 80\% and token usage by 90\%. The resulting PRMs exhibit strong step-quality assessment capabilities and yield consistent improvements in BoN inference across various mathematical benchmarks. These results demonstrate that CPMI provides a practical and resource-efficient alternative to MC-based supervision, particularly in settings with limited computational budgets.

\section*{Limitations}
We validate our methods only during the inference stage. While reward or verifier models are typically employed within online reinforcement learning frameworks, our CPMI-based approach could naturally be extended to such training settings. In this work, we focus on its efficiency and discriminative capacity under offline evaluation. Another limitation lies in the scale of our trajectory data. We collect around 80K process-level samples, which is relatively small compared with recent large-scale PRM datasets. Future work can explore larger and more diverse trajectories, as well as varying PRM model sizes beyond the current Qwen3-4B configuration, to better understand scalability and robustness across different model capacities.

\section*{Acknowledgments}
This work is in part supported by the National Research Foundation of Korea (NRF, RS-2024-00451435(20\%), RS-2024-00413957(20\%)), Institute of Information \& communications Technology Planning \& Evaluation (IITP, RS-2025-02305453(15\%), RS-2025-02273157(15\%), RS-2025-25442149(15\%) RS-2021-II211343(15\%)) grant funded by the Ministry of Science and ICT (MSIT), Institute of New Media and Communications(INMAC), and the BK21 FOUR program of the Education, Artificial Intelligence Graduate School Program (Seoul National University), and Research Program for Future ICT Pioneers, Seoul National University in 2026.

\newpage
\bibliography{custom}

@inproceedings{chainofthought,
  author    = {Jason Wei and Xuezhi Wang and Dale Schuurmans and Maarten Bosma and Brian Ichter and Fei Xia and Ed H. Chi and Quoc V. Le and Denny Zhou},
  title     = {Chain-of-Thought Prompting Elicits Reasoning in Large Language Models},
  booktitle = {Advances in Neural Information Processing Systems (NeurIPS) 35},
  pages     = {24824--24837},
  year      = {2022},
  url       = {https://arxiv.org/abs/2201.11903},
  bibsource = {arXiv:2201.11903v6}
}

@article{selfconsistency,
  author    = {Xuezhi Wang and Jason Wei and Dale Schuurmans and Quoc V. Le and Ed H. Chi and Sharan Narang and Aakanksha Chowdhery and Denny Zhou},
  title     = {Self-Consistency Improves Chain of Thought Reasoning in Language Models},
  journal   = {CoRR},
  volume    = {abs/2203.11171},
  year      = {2022},
  url       = {https://arxiv.org/abs/2203.11171},
  bibsource = {arXiv:2203.11171v1}
}

@inproceedings{mathshepherd,
  author    = {Peiyi Wang and Lei Li and Zhihong Shao and Runxin Xu and Damai Dai and Yifei Li and Deli Chen and Yu Wu and Zhifang Sui},
  title     = {Math-Shepherd: Verify and Reinforce LLMs Step-by-step without Human Annotations},
  booktitle = {Proceedings of the 62nd Annual Meeting of the Association for Computational Linguistics (Volume 1: Long Papers)},
  pages     = {9426--9439},
  year      = {2024},
  month     = {August},
  address   = {Bangkok, Thailand},
  publisher = {Association for Computational Linguistics},
  doi       = {10.18653/v1/2024.acl-long.510},
  url       = {https://aclanthology.org/2024.acl-long.510},
}

@article{qwenprm,
  author    = {Zhenru Zhang and Chujie Zheng and Yangzhen Wu and Beichen Zhang and Runji Lin and Bowen Yu and Dayiheng Liu and Jingren Zhou and Junyang Lin},
  title     = {The Lessons of Developing Process Reward Models in Mathematical Reasoning},
  journal   = {CoRR},
  volume    = {abs/2501.07301},
  year      = {2025},
  url       = {https://arxiv.org/abs/2501.07301},
  note      = {arXiv preprint arXiv:2501.07301v2},
}

@inproceedings{processbench,
  title     = {ProcessBench: Identifying Process Errors in Mathematical Reasoning},
  author    = {Zheng, Chen and others},
  booktitle = {Proceedings of the 63rd Annual Meeting of the Association for Computational Linguistics (ACL)},
  year      = {2025},
  url       = {https://aclanthology.org/2025.acl-long.50/}
}

@article{epic,
  title   = {An Efficient and Precise Training Data Construction Framework for Process-supervised Reward Model in Mathematical Reasoning},
  author  = {Sun, Weiqi and others},
  journal = {arXiv preprint arXiv:2503.02382},
  year    = {2025},
  url     = {https://arxiv.org/abs/2503.02382}
}

@article{stepwiser,
  title   = {StepWiser: Stepwise Generative Judges for Wiser Reasoning},
  author  = {Xiong, Wei and Zhao, Wenting and Yuan, Weizhe and Golovneva, Olga and Zhang, Tong and Weston, Jason and Sukhbaatar, Sainbayar},
  journal = {arXiv preprint arXiv:2508.19229},
  year    = {2025},
  url     = {https://arxiv.org/abs/2508.19229}
}

@misc{rewardingprocess,
      title={Rewarding Progress: Scaling Automated Process Verifiers for LLM Reasoning}, 
      author={Amrith Setlur and Chirag Nagpal and Adam Fisch and Xinyang Geng and Jacob Eisenstein and Rishabh Agarwal and Alekh Agarwal and Jonathan Berant and Aviral Kumar},
      year={2024},
      eprint={2410.08146},
      archivePrefix={arXiv},
      primaryClass={cs.LG},
      url={https://arxiv.org/abs/2410.08146}, 
}

@article{alphamath,
  title   = {AlphaMath Almost Zero: Process Supervision without Process},
  author  = {Chen, Guoxin and Liao, Minpeng and Li, Chengxi and Fan, Kai},
  journal = {arXiv preprint arXiv:2405.03553},
  year    = {2024},
  url     = {https://arxiv.org/abs/2405.03553}
}

@article{pot,
  title   = {Program of Thoughts Prompting: Disentangling Computation from Reasoning for Numerical Reasoning Tasks},
  author  = {Chen, Wenhu and others},
  journal = {arXiv preprint arXiv:2211.12588},
  year    = {2022},
  url     = {https://arxiv.org/abs/2211.12588}
}

@misc{pal,
      title={PAL: Program-aided Language Models}, 
      author={Luyu Gao and Aman Madaan and Shuyan Zhou and Uri Alon and Pengfei Liu and Yiming Yang and Jamie Callan and Graham Neubig},
      year={2023},
      eprint={2211.10435},
      archivePrefix={arXiv},
      primaryClass={cs.CL},
      url={https://arxiv.org/abs/2211.10435}, 
}

@inproceedings{decompose,
  title     = {Least-to-Most Prompting Enables Complex Reasoning in Large Language Models},
  author    = {Zhou, Denny and Sch{\"a}rli, Nathanael and Hou, Le and Wei, Jason and Scales, Nathan and Wang, Xuezhi and Schuurmans, Dale and Cui, Claire and Bousquet, Olivier and Le, Quoc and Chi, Ed H.},
  booktitle = {International Conference on Learning Representations (ICLR)},
  year      = {2023},
  url       = {https://openreview.net/forum?id=WZH7099tgfM}
}

@article{pure,
  title   = {Stop Summation: Min-Form Credit Assignment Is All Process Reward Model Needs for Reasoning},
  author  = {Cheng, Jie and Qiao, Ruixi and Li, Lijun and Guo, Chao and Wang, Junle and Xiong, Gang and Lv, Yisheng and Wang, Fei-Yue},
  journal = {arXiv preprint arXiv:2504.15275},
  year    = {2025},
  note    = {PURE: PRM-supervised reinforcement fine-tuning for reasoning},
  url     = {https://arxiv.org/abs/2504.15275}
}

@techreport{qwen3tech,
  title        = {Qwen3 Technical Report},
  author       = {Qwen Team},
  institution  = {Qwen Model Team, Qwen Language Model},
  number       = {arXiv:2505.09388},
  year         = {2025},
  url          = {https://arxiv.org/pdf/2505.09388},
  note         = {Open weight large language model family (0.6B–235B parameters) with thinking/non-thinking modes, multilingual capabilities and public release under Apache 2.0}
}

@article{tdlambda,
  author    = {Richard S. Sutton},
  title     = {Learning to Predict by the Methods of Temporal Differences},
  journal   = {Machine Learning},
  volume    = {3},
  number    = {1},
  pages     = {9--44},
  year      = {1988},
  doi       = {10.1007/BF00115009}
}

@inproceedings{prmbench,
  title        = "{PRMB}ench: A Fine-grained and Challenging Benchmark for Process-Level Reward Models",
  author       = "Song, Mingyang  and Su, Zhaochen  and Qu, Xiaoye  and Zhou, Jiawei  and Cheng, Yu",
  editor       = "Che, Wanxiang  and Nabende, Joyce  and Shutova, Ekaterina  and Pilehvar, Mohammad Taher",
  booktitle    = "Proceedings of the 63rd Annual Meeting of the Association for Computational Linguistics (Volume 1: Long Papers)",
  year         = "2025",
  address      = "Vienna, Austria",
  publisher    = "Association for Computational Linguistics",
  url          = "https://aclanthology.org/2025.acl-long.1230/",
  doi          = "10.18653/v1/2025.acl-long.1230",
}

@article{gsm8k,
  title  = {Training Verifiers to Solve Math Word Problems},
  author = {Cobbe, Karl and Kosaraju, Vineet and Bavarian, Mohammad and Chen, Mark and Jun, Heewoo and Kaiser, Łukasz and Plappert, Matthias and Tworek, Jerry and Hilton, Jacob and Nakano, Reiichiro and Hesse, Christopher and Schulman, John},
  journal= {arXiv preprint arXiv:2110.14168},
  year   = {2021},
  url    = {https://doi.org/10.48550/arXiv.2110.14168}
}

@misc{math,
      title={Measuring Mathematical Problem Solving With the MATH Dataset}, 
      author={Dan Hendrycks and Collin Burns and Saurav Kadavath and Akul Arora and Steven Basart and Eric Tang and Dawn Song and Jacob Steinhardt},
      year={2021},
      eprint={2103.03874},
      archivePrefix={arXiv},
      primaryClass={cs.LG},
      url={https://arxiv.org/abs/2103.03874}, 
}

@article{mmlu,
  title  = {Measuring Massive Multitask Language Understanding},
  author= {Hendrycks, Dan and Burns, Collin and Basart, Steven and Zou, Andy and Mazeika, Mantas and Song, Dawn and Steinhardt, Jacob},
  journal= {Proceedings of the International Conference on Learning Representations (ICLR)},
  year   = {2021},
  url    = {https://doi.org/10.48550/arXiv.2009.03300}
}

@inproceedings{omni,
  title     = {Omni-MATH: A Universal Olympiad Level Mathematic Benchmark for Large Language Models},
  author    = {Gao, Bofei and Song, Feifan and Yang, Zhe and Cai, Zefan and Miao, Yibo and others},
  booktitle = {Proceedings of the International Conference on Learning Representations (ICLR)},
  year      = {2024},
  url       = {https://arxiv.org/abs/2410.07985}
}

@misc{rlp,
  title        = {RLP: Reinforcement as a Pretraining Objective},
  author       = {Hatamizadeh, Ali and Akter, Syeda Nahida and Prabhumoye, Shrimai and Kautz, Jan and Patwary, Mostofa and Shoeybi, Mohammad and Catanzaro, Bryan and Choi, Yejin},
  year         = {2025},
  howpublished = {arXiv preprint arXiv:2510.01265},
  url          = {https://arxiv.org/abs/2510.01265}
}

@misc{nover,
  title={NOVER: Incentive Training for Language Models via Verifier-Free Reinforcement Learning},
  author={Wei Liu and Siya Qi and Xinyu Wang and Chen Qian and Yali Du and Yulan He},
  year={2025},
  url = {https://arxiv.org/abs/2505.16022},
  doi = {10.48550/arXiv.2505.16022}
}

@misc{rlif,
  title={Learning to Reason without External Rewards},
  author       = {Xuandong Zhao and Zhewei Kang and Aosong Feng and Sergey Levine and Dawn Song},
  url          = {https://arxiv.org/abs/2505.19590},
  doi          = {10.48550/arXiv.2505.19590},
  year={2025}
}

@inproceedings{KHS,
  author    = {Jeong, Geunyeong and Sun, Juoh and Kim, Harksoo},
  title     = {Watch Your Step: A Fine-Grained Evaluation Framework for Multi-hop Knowledge Editing in Large Language Models},
  year      = {2025},
  isbn      = {9798400720406},
  publisher = {Association for Computing Machinery},
  address   = {New York, NY, USA},
  url       = {https://doi.org/10.1145/3746252.3760840},
  doi       = {10.1145/3746252.3760840},
  booktitle = {Proceedings of the 34th ACM International Conference on Information and Knowledge Management},
  pages     = {4842--4846},
  location  = {Seoul, Republic of Korea},
  series    = {CIKM '25}
}

@article{OBK, title={HLMEA: Unsupervised Entity Alignment Based on Hybrid Language Models}, volume={39}, url={https://ojs.aaai.org/index.php/AAAI/article/view/33294}, DOI={10.1609/aaai.v39i11.33294}, 
number={11}, journal={Proceedings of the AAAI Conference on Artificial Intelligence}, author={Jin, Xiongnan and Wang, Zhilin and Chen, Jinpeng and Yang, Liu and Oh, Byungkook and Hwang, Seung-won and Li, Jianqiang}, year={2025}, month={Apr.}, pages={11888-11896} }

\clearpage
\appendix

\section{Practical Implementations}
\label{app:imple}
In practice, LLMs are highly sensitive to prompt phrasing, which can bias the estimated step contributions and increase variance. To obtain stable and reliable reward estimates, we incorporate two key design choices: (\textit{i}) Prompt diversification and the construction of hard negative answers, (\textit{ii}) length normalization when computing the CPMI. 

\paragraph{Prompt diversification.} 
To mitigate the sensitivity of prompt wording, we employ $K$ distinct prompt templates. Each template presents the problem statement $q$ with minor structural or formatting variations. Averaging across these templates reduces prompt-specific variance and yields more stable step-level rewards. We use the four prompts shown in Figure~\ref{app-fig:promptdiv}, which convey the same instruction but are phrased in slightly different ways. 

We use the prompt templates in Figure~\ref{app-fig:neg}, varying the number of samples $M$.

\begin{figure}[h]
    \centering
    \tcbset{
        promptbox/.style={
            colback=white,
            colframe=black!70, 
            boxrule=0.8pt,
            arc=2pt, 
            left=2pt, right=2pt, top=2pt, bottom=2pt,
            fonttitle=\bfseries\small,
            title={\strut Prompt \#\thetcbrasternum}, 
            fontupper=\small\ttfamily, 
            halign=left,
            breakable 
        }
    }
    \begin{tcbraster}[raster columns=1, raster equal height, raster column skip=1em, raster row skip=1em]
    
    \begin{tcolorbox}[promptbox]
You are a careful math solver. Follow the steps methodically. Keep each step concise.\textbackslash n
At the end, output exactly one line in the format:\textbackslash n
The answer is: <final answer>\textbackslash n\textbackslash n
Problem:\textbackslash n\{q\}\textbackslash n
Solution: Let's think step by step.\textbackslash n
    \end{tcolorbox}
    \begin{tcolorbox}[promptbox]
Solve the problem with numbered steps. Be precise. Finally print exactly:\textbackslash n
The answer is: <final answer>\textbackslash n
If numeric, use plain digits only (no punctuation).\textbackslash n\textbackslash n
Problem:\textbackslash n\{q\}\textbackslash n
Solution (step-by-step):\textbackslash n
    \end{tcolorbox}
    \begin{tcolorbox}[promptbox]
Work through the solution briefly, then verify and conclude. Conclude with exactly:\textbackslash n
The answer is: <final answer>\textbackslash n\textbackslash n
Problem:\textbackslash n\{q\}\textbackslash n
Solution: Let's proceed carefully.\textbackslash n
    \end{tcolorbox}
    \begin{tcolorbox}[promptbox]
You are solving a math problem. Compute step by step. End with exactly:\textbackslash n 
The answer is: <final answer>\textbackslash n\textbackslash n
Problem:\textbackslash n\{q\}\textbackslash n
Solution: Let's think step by step.\textbackslash n
    \end{tcolorbox}
    \end{tcbraster}
    \caption{The four prompts used for diversification.}
    \label{app-fig:promptdiv}
    \vspace{-0.2in}
\end{figure}

\begin{figure}[h]
    \centering
    \tcbset{
        promptbox/.style={
            colback=white,
            colframe=black!70, 
            boxrule=0.8pt,
            arc=2pt, 
            left=4pt, right=4pt, top=4pt, bottom=4pt, 
            fonttitle=\bfseries\small,
            fontupper=\small\ttfamily, 
            halign=left,
        }
    }
\begin{tcolorbox}[promptbox, title={Hard-Negative Generation Prompt Template}]
    <Base Prompt containing the Problem Question>

(Gold answer: <Gold Truth Value>)

Generate several plausible but incorrect answers to the given question. Identify any wrong step and take that step's wrong computed result or the mistaken number used in it as a hard-negative example. Try to include all \textbf{earlier steps' wrong intermediate outputs} if possible. If all steps seem correct, return a plausible wrong number ($\pm$1, $\pm$2, or $\pm$10\%).

Output exactly one short line after 'The answer is:' no units or words. \textbackslash n

The answer is: 
    \end{tcolorbox}
    \caption{Prompt for generating hard-negative answers.}
    \label{app-fig:neg}
    \vspace{-0.2in}
\end{figure}

\paragraph{Length Normalization.} 
When computing the pointwise mutual information in Eq.~\ref{eq:cpmi-marg}, we compute the log-probability as $\log p_{\theta}(a \mid X) = \frac{1}{L} \sum_{l=1}^{L} \log p_{\theta}(a_l \mid a_{<l}, X)$, applying length normalization to mitigate the influence of varying sequence lengths.

\section{Statistics of CPMI80k}
\begin{table}[h]
    \centering
    \small
    \begin{tabular}{l|c}
    \toprule
    \textbf{Statistic} & \textbf{Ratio (\%) / Count} \\ 
    \midrule
    Total samples & 80,000 \\
    Task Source (GSM8K : MATH) & $1 : 1$ \\
    Composition (mixed : pure) & $7 : 3$ \\
    \bottomrule
    \end{tabular}
    \caption{Statistics of the \textsc{CPMI-80k} dataset. All values are reported as counts or ratios.}
    \label{app-tab:cpmi80k}
    \vspace{-0.1in}
\end{table}

When constructing the 80k-sample from the Math-Shepherd corpus, we sample an equal number of problems from GSM8K and MATH to ensure coverage of both easier and more challenging reasoning tasks. We exclude extremely difficult problems, as PRMs fail to extract meaningful learning signals when all step-wise rewards collapse to zero. Additionally, we intentionally include a higher proportion of \emph{mixed} trajectories, containing both positive ($+$) and negative ($-$) step labels, than \emph{pure} trajectories with a single label type, as mixed trajectories provide richer supervisory signals for training step-level reward models.

\begin{table*}[t]
\centering
\small
\begin{tabular}{
  @{}
  l
  *{3}{S[table-format=2.1]}
  *{3}{S[table-format=2.1]}
  *{3}{S[table-format=2.1]}
  *{3}{S[table-format=2.1]}
  >{\columncolor{effcol}}S[table-format=2.1]
  @{}
}
\toprule
& \multicolumn{3}{c}{\textbf{GSM8K}}
& \multicolumn{3}{c}{\textbf{Math}}
& \multicolumn{3}{c}{\textbf{Olympiad}}
& \multicolumn{3}{c}{\textbf{Omni}}
& \textbf{Average} \\
\cmidrule(lr){2-4} \cmidrule(lr){5-7} \cmidrule(lr){8-10} \cmidrule(lr){11-13}
\textbf{Reward Type} &
\multicolumn{1}{c}{\textbf{Err}} &
\multicolumn{1}{c}{\textbf{Corr}} &
\multicolumn{1}{c}{\textbf{F1}} &
\multicolumn{1}{c}{\textbf{Err}} &
\multicolumn{1}{c}{\textbf{Corr}} &
\multicolumn{1}{c}{\textbf{F1}} &
\multicolumn{1}{c}{\textbf{Err}} &
\multicolumn{1}{c}{\textbf{Corr}} &
\multicolumn{1}{c}{\textbf{F1}} &
\multicolumn{1}{c}{\textbf{Err}} &
\multicolumn{1}{c}{\textbf{Corr}} &
\multicolumn{1}{c}{\textbf{F1}} & 
\textbf{F1} \\
\midrule
MC & 22.2 & 85.0 & 35.2 & 25.9 & 76.8 & 38.8 & 10.7 & 54.3 & 17.9 & 11.5 & 53.9 & 18.9 & \cellcolor{effcol} 27.7 \\
PAV & 35.7 & 93.8 & 51.8 & 32.2 & 68.5 & \textbf{43.8} & 17.5 & 64.3 & 27.6 & 14.5 & 57.3 & 23.1 & \cellcolor{effcol} 36.6\\
\midrule
CPMI & 35.7 & 79.8 & 49.4 & 29.0 & 62.8 & 39.6 & 15.9 & 53.7 & 24.5 & 17.5 & 42.3 & 24.8 & \cellcolor{effcol} 34.6 \\
CPMI\_Merge & 37.2 & 86.5 & \textbf{52.0} & 31.0 & 59.6 & 40.8 & 20.1 & 49.9 & \textbf{28.7} & 18.4 & 41.9 & \textbf{25.6} & \cellcolor{effcol} \textbf{36.8} \\
\midrule
Rand\_Merge & 13.5 & 55.4 & 21.7 & 16.5 & 50.5 & 24.9 & 7.1 & 43.4 & 12.2 & 9.6 & 46.1 & 15.9 & \cellcolor{effcol} 18.7 \\
\midrule
MS* & 32.4 & 91.7 & 47.9 & 18.2 & 69.2 & 33.9 & 15.1 & 77.1 & 24.8 & 12.4 & 72.3 & 19.8 & \cellcolor{effcol} 31.5 \\
RLHFlow* & 33.8 & 99.0 & 50.4 & 21.4 & 72.2 & 33.4 &  8.2 & 43.1 & 13.8 &  9.6 & 45.2 & 15.8 & \cellcolor{effcol} 28.4 \\
\bottomrule
\end{tabular}
\caption{Detailed Performance by reward type on ProcessBench. The last two rows marked with * correspond to the reported performance in \citet{mathshepherd}. Bold values indicate the best F1 score in each column.}
\label{app-tab:pb}
\end{table*}

\begin{table*}[t]
\centering
\setlength{\tabcolsep}{6pt}
\begin{tabular}{lcc cccc ccc c}
\toprule
& \multicolumn{2}{c}{\textbf{Simplicity}} 
& \multicolumn{4}{c}{\textbf{Soundness}} 
& \multicolumn{3}{c}{\textbf{Sensitivity}} 
& \textbf{Overall} \\
\cmidrule(lr){2-3}\cmidrule(lr){4-7}\cmidrule(lr){8-10}
\textbf{Reward Type} & NR & NCL & ES & SC & DC & CI & PS & DR & MS & \\
\midrule
MC          & 37.56 & 44.20 & 43.18 & 26.84 & 36.72 & 40.94 & 42.28 & 42.30 & 20.61 & 38.8 \\
PAV         & 45.69 & 48.62 & 55.86 & 39.09 & 46.96 & 56.32 & 50.17 & 53.18 & 38.25 & 49.64 \\
\midrule
CPMI        & 50.22 & 54.89 & 63.76 & 56.99 & 56.58 & 63.52 & 57.70 & 62.02 & \textbf{47.06} & 58.75 \\
CPMI\_Merge & \textbf{51.27} & \textbf{58.02} & \textbf{65.99} & \textbf{58.96} & \textbf{58.70} & \textbf{65.53} & \textbf{59.28} & \textbf{63.34} & 46.89 & \textbf{60.67} \\
Rand\_Merge & 20.14 & 35.15 & 26.40 & 19.26 & 21.11 & 19.79 & 25.71 & 23.24 &  7.00 & 23.20 \\
\bottomrule
\end{tabular}
\caption{PRMBench results across reward designs. Bold values indicate the best PRMScore.}
\label{app-tab:prmbench}
\vspace{-0.1in}
\end{table*}

\section{Fine-Grained Process-Level Evaluation}
\label{app-sec:detailed}
We provide extended results on ProcessBench and PRMBench in Table~\ref{app-tab:pb} and Table~\ref{app-tab:prmbench}, respectively. Table~\ref{app-tab:pb} reports three metrics for each dataset—error accuracy (Err), correct accuracy (Corr), and F1—along with their mean F1 in the \textbf{Average} column. We observe that the performance gaps between MC, PAV, and our CPMI-based methods are more pronounced in challenging out-of-domain evaluations such as Olympiad and Omni, highlighting CPMI's improved generalization to complex reasoning tasks. 

Table~\ref{app-tab:prmbench} presents fine-grained performance according to the nine evaluation dimensions introduced in~\citet{prmbench}: Non-Redundancy (NR), Non-Circular Logic (NCL), Empirical Soundness (ES), Step Consistency (SC), Domain Consistency (DC), Confidence Invariance (CI), Prerequisite Sensitivity (PS), Deception Resistance (DR), and Multi-Solution Consistency (MS). These results demonstrate that CPMI enhances the model's capability to distinguish diverse failure modes in step-level reasoning, offering more reliable reward signals across a wide range of logical and mathematical phenomena.

\section{Robustness of CPMI}
\label{app:robustness}

\subsection{Robustness Across Labeling Model Capacities}
We further evaluate CPMI on a different base model family, focusing on whether the proposed reward construction remains effective even when the labeling model is much smaller. 

Table~\ref{app-tab:small_robustness} shows that CPMI remains consistently effective even with compact labeling models from two different families, \texttt{Qwen2.5-1.5B} and \texttt{Llama3.2-1B}. In both cases, replacing MC labeling with CPMI-based labeling improves the step-level discrimination metric (ROC-AUC) and yields large gains on process-level benchmarks, especially on ProcessBench and PRMBench. Importantly, these quality improvements are achieved while substantially reducing reward construction cost, requiring 65--73\% less labeling time and 82\% fewer generated tokens.

These results suggest that although stronger pretrained models can provide cleaner intrinsic signals, CPMI does not depend on a single high-capacity model regime. Instead, it offers a robust and cost-effective reward proxy even with lightweight labeling backbones, making the overall pipeline more practical for resource-constrained settings.

\begin{table*}[t]
\centering
\small
\renewcommand{\arraystretch}{1.08}
\setlength{\tabcolsep}{4pt}
\begin{tabular}{llcccccc}
\toprule
\textbf{Labeler} & \textbf{Reward} & \textbf{ROC-AUC} $\uparrow$ & \textbf{PB} $\uparrow$ & \textbf{PRMB} $\uparrow$ & \textbf{MATH500} $\uparrow$ & $\Delta$\textbf{Time} & $\Delta$\textbf{Token} \\
\midrule
\multirow{2}{*}{Qwen2.5-1.5B}
& MC         & 0.72 & 28.4 & 16.2 & 41.6 & 0\%   & 0\% \\
& CPMI-based & \textbf{0.75} & \textbf{36.1} & \textbf{51.8} & \textbf{45.8} & \textbf{-65\%} & \textbf{-82\%} \\
\midrule
\multirow{2}{*}{Llama3.2-1B}
& MC         & 0.73 & 29.5 & 18.6 & \textbf{48.0} & 0\%   & 0\% \\
& CPMI-based & \textbf{0.76} & \textbf{37.0} & \textbf{59.1} & 47.6 & \textbf{-73\%} & \textbf{-82\%} \\
\bottomrule
\end{tabular}
\caption{Robustness of CPMI across smaller labeling models from different families.}
\label{app-tab:small_robustness}
\end{table*}

\subsection{Generalization to Logical Reasoning}
To further validate the robustness of our reward signal beyond mathematical reasoning, we additionally evaluate PRMs trained with different reward constructions on out-of-distribution logical reasoning benchmarks. Following the main evaluation setup, we use Qwen3-4B-Base under zero-shot Best-of-8 inference.

As shown in Table~\ref{app-tab:logic_ood}, both CPMI and CPMI\_Merge consistently outperform MC across all three benchmarks, including FOLIO, LogiQA 2.0, and LogicNLI. These results indicate that the benefits of CPMI are not limited to in-domain math reasoning, but also transfer to logical reasoning tasks that require process-level verification.

\begin{table}[t]
\centering
\small
\renewcommand{\arraystretch}{1.08}
\setlength{\tabcolsep}{4pt}
\begin{tabular}{lccc}
\toprule
\textbf{Reward Type} & \textbf{FOLIO} $\uparrow$ & \textbf{LogiQA 2.0} $\uparrow$ & \textbf{LogicNLI} $\uparrow$ \\
\midrule
MC           & 54.19 & 33.39 & 25.4 \\
CPMI         & \textbf{59.61} & \textbf{37.71} & \textbf{32.1} \\
CPMI\_Merge  & 58.62 & 35.66 & 30.8 \\
Rand\_Merge  & 42.86 & 24.54 & 24.5 \\
\bottomrule
\end{tabular}
\caption{Generalization to out-of-distribution logical reasoning benchmarks under zero-shot BoN@8.}
\label{app-tab:logic_ood}
\end{table}

\subsection{Robustness Across Training Objectives}
Our core contribution lies in the labeling signal and dataset construction, which are largely agnostic to the downstream training objective. We use BCE in the main experiments to match the Math-Shepherd training setup as closely as possible, ensuring a fair comparison with MC-based baselines. Since CPMI provides continuous step-wise supervision after normalization, it can naturally be paired with objectives beyond BCE.

To verify this, we additionally train PRMs with two alternative objectives: mean squared error (MSE), which treats reward prediction as a regression problem, and pairwise ranking loss (PQM), which emphasizes relative ordering between steps. As shown in Table~\ref{app-tab:loss_robustness}, CPMI remains consistently competitive across all training objectives and continues to outperform MC in PRMBench and MATH500 under both BCE and MSE.

Interestingly, PQM+MC achieves the highest ProcessBench score, suggesting that pairwise ranking can act as a highly localized error detector. However, this gain does not transfer to broader process-level quality measures, where CPMI-based supervision remains substantially stronger, especially on PRMBench. Overall, these results support our claim that the advantage of CPMI primarily comes from the reward signal itself rather than from a specific loss design.

\begin{table}[t]
\centering
\small
\renewcommand{\arraystretch}{1.08}
\setlength{\tabcolsep}{3pt}
\begin{tabular}{llccc}
\toprule
\textbf{Loss} & \textbf{Reward} & \textbf{ProcessBench} & \textbf{PRMBench} & \textbf{MATH500} \\
\midrule
\multirow{2}{*}{BCE}
& MC   & 27.7 & 38.8 & 45.4 \\
& CPMI & \textbf{34.6} & \textbf{58.8} & \textbf{48.2} \\
\midrule
\multirow{2}{*}{PQM}
& MC   & \textbf{37.9} & 45.4 & 44.0 \\
& CPMI & 35.6 & \textbf{56.7} & \textbf{46.2} \\
\midrule
\multirow{2}{*}{MSE}
& MC   & 31.0 & 43.1 & 45.4 \\
& CPMI & \textbf{35.5} & \textbf{59.9} & \textbf{47.2} \\
\bottomrule
\end{tabular}
\caption{Robustness of CPMI across different PRM training objectives.}
\label{app-tab:loss_robustness}
\end{table}

\section{Ablation Experiments}
We conduct a series of ablation studies to evaluate the robustness and generalizability of our CPMI reward design. First, we vary the number of hard-negative targets to analyze how contrastive signals influence performance (Section~\ref{app:hardneg}). We then examine the efficiency and performance–cost trade-offs of different step at which CPMI replaces MC used in the CPMI\_Merge variants (Section~\ref{app:switchat}). Finally, we demonstrate the flexibility of CPMI by merging it with PAV-based baselines (Section~\ref{app:pavmerge}).

\subsection{Ablations on the Hard-Negative Term}
\label{app:hardneg} 
\begin{figure*}[h]
\begin{center}
\centerline{\includegraphics[width=\textwidth, trim=0cm 3.5cm 0cm 4cm, clip]{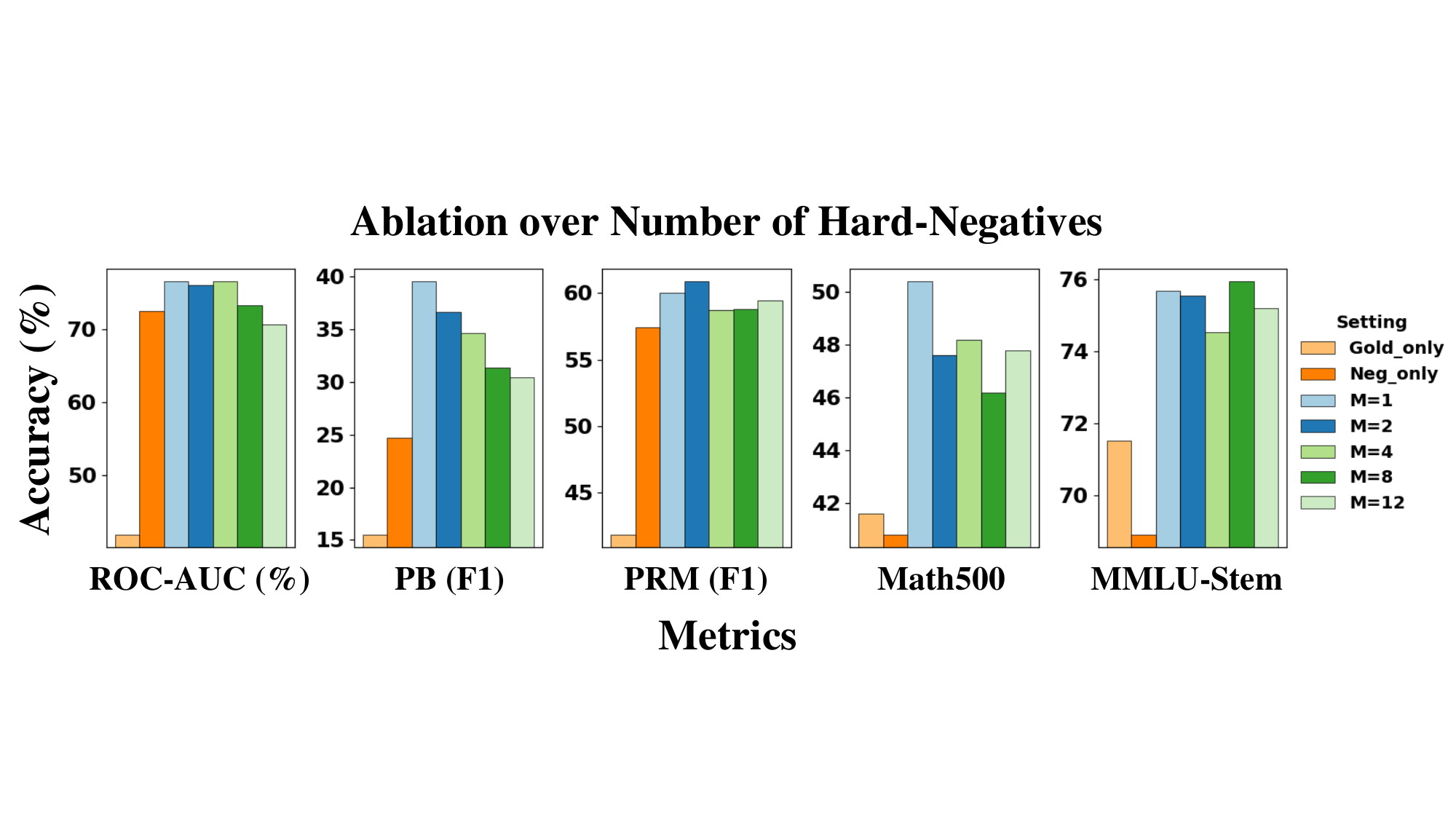}}
\caption{Ablation results varying the number of hard-negative targets $M$ in the CPMI objective. We report both process-level metrics (ROC-AUC, PB, PRM) and downstream performance (Math, MMLU).}
\label{app-fig:mabl}
\end{center}
\vspace{-0.2in}
\end{figure*}

Figure~\ref{app-fig:mabl} varies the number of hard-negative answers $M$ used in the CPMI objective. Transitioning from $M=0$ (Gold\_only) or using only the negative term (Neg\_only) to any setting with at least one negative sample ($M \ge 1$) yields substantial improvements across all evaluation metrics, as mentioned in Section~\ref{sec:contrastive}. 
Interestingly, contrary to the expectation that increasing $M$ would yield more stable performance by reducing variance, $M=1$ achieves the strongest Math accuracy and competitive performance on the other benchmarks. We attribute this behavior to a trade-off between signal strength and averaging: when $M=1$, the CPMI reward focuses on the single hardest negative, providing a strong contrastive signal, whereas larger $M$ values average over easier or noisier negatives, diluting the gradient despite reducing variance.

At the same time, the differences among $M \in \{1,2,4\}$ are relatively small (within a few points across benchmarks), suggesting that CPMI is not overly sensitive to the precise choice of $M$ once at least one hard negative is included. In our main experiments, we therefore fix $M=4$ as a default, which offers a stable estimate of the negative term and reduces sensitivity to the particular sampled negatives, while maintaining performance close to the best ablation setting. Overall, these results indicate that the key factor is the presence of explicit contrastive negatives, and that moderate values of $M$ (e.g., $M=1$–$4$) strike a reasonable balance between robustness and computational cost.

\subsection{Ablations on Different Merge Points}
\label{app:switchat}
Table~\ref{app-tab:eff_merge} reports the resulting performance across four metrics: step-level discriminative ability measured by ROC-AUC (\textbf{AUC}), average F1 on ProcessBench (\textbf{PB}), the overall PRMScore from PRMBench (\textbf{PRM}), and MATH500 BoN@8 accuracy using PRM-weighted trajectory selection (\textbf{MATH}). 
We use MC reward only at the first step as an initializer, we did ablation experiments of diverging the merge index to figure out the sweet spot for performance cost efficiency trade-off. We additionally report the computational cost of annotating step-level datasets. The relative efficiency score (\textbf{RelEff}) are computed using Eq.~\eqref{eq:rel-eff}. As shown in Table~\ref{app-tab:eff_merge}, merging at index~1 yields the strongest overall performance across benchmarks while preserving a high level of efficiency, achieving relative efficiency gains in the range of 3.4$\times$ to 5.2$\times$. Although later merge indices maintain competitive performance, their efficiency declines substantially due to the increased computational cost associated with relying on MC rewards for more steps. These results indicate that applying MC only at the earliest step and transitioning to CPMI immediately after provides the most favorable balance between quality and cost.

\begin{table*}[t]
\centering
\resizebox{\textwidth}{!}{
\begin{tabular}{lccccccccccccc}
\toprule
 &  &  & \multicolumn{4}{c}{\textbf{Model Quality}} 
& \multicolumn{3}{c}{\textbf{Computation Cost (Ratio)}} 
& \multicolumn{4}{c}{\textbf{RelEff ($\times$)}} \\
\cmidrule(lr){4-7} \cmidrule(lr){8-10} \cmidrule(lr){11-14}
\textbf{Merge Index} & \textbf{MC(\%)} & \textbf{CPMI(\%)} & ROC-AUC & PB & PRMB & MATH & Time & ratio & gen\_tokens & AUC & PB & PRMB & MATH \\
\midrule
CPMI & 0   & 100 & 0.765 & 34.6 & 58.8 & 48.2 & 38{,}603  & 0.16 & 1{,}579{,}148 & \textbf{6.34} & \textbf{7.85} & \textbf{9.53} & \textbf{6.68} \\
\midrule
CPMI\_Merge1 & 16.6 & 83.4 & 0.766 & \textbf{36.8} & \textbf{60.7} & 49.4 & 72{,}553  & 0.30 & 16{,}420{,}803 & 3.38 & 4.45 & 5.24 & 3.64 \\
CPMI\_Merge2 & 33.2 & 66.8 & 0.784 & 35.8 & 58.9 & 47.0 & 106{,}452 & 0.44 & 31{,}239{,}824 & 2.36 & 2.95 & 3.47 & 2.36 \\
CPMI\_Merge3 & 48.0 & 52.0 & \textbf{0.804} & 36.6 & 56.7 & \textbf{56.6} & 136{,}579 & 0.56 & 44{,}410{,}422 & 1.88 & 2.35 & 2.60 & 2.22 \\
\midrule
MC  & 100 & 0   & 0.759 & 27.7 & 38.8 & 45.4 & 242{,}930 & 1.00 & 90{,}902{,}162 & 1.00 & 1.00 & 1.00 & 1.00 \\
\bottomrule
\end{tabular}
}
\caption{Efficiency analysis of CPMI merging variants across quality, cost, and cost-effectiveness metrics.}
\label{app-tab:eff_merge}
\end{table*}

\subsection{Initialized with PAV}
\label{app:pavmerge}
Since this initialization strategy is compatible with other labeling methods as well, we additionally evaluate a variant that uses the PAV reward for the first step and applies CPMI to all subsequent steps. We assess performance using four metrics, the same metrics used in Section.~\ref{app:switchat}. As shown in Table~\ref{app-tab:pav_merg}, initializing CPMI with PAV yields stable behavior and achieves a favorable performance–efficiency trade-off, mirroring the trends observed when MC is used as the initializer.

\begin{table*}[t]
\centering
\setlength{\tabcolsep}{4pt}
\resizebox{\textwidth}{!}{%
\begin{tabular}{lrrrrrrrrrr}
\toprule
& \multicolumn{4}{c}{\textbf{Model Quality}} &
\multicolumn{2}{c}{\textbf{Computation Cost (Ratio)}} &
\multicolumn{4}{c}{\textbf{RelEff ($\mathbf{\times}$})} \\
\cmidrule(lr){2-5}\cmidrule(lr){6-7}\cmidrule(lr){8-11}
\textbf{Reward Type} & \textbf{AUC} & \textbf{PB} & \textbf{PRMB} & \textbf{MATH} & \textbf{Time} & \textbf{Token} & \textbf{AUC} & \textbf{PB} & \textbf{PRMB} & \textbf{MATH} \\
\midrule
PAV & 0.757 & 36.6 & 49.6 & 47.2 & 1.00 & 1.00 & 1.00 & 1.00 & 1.00 & 1.00 \\
\midrule
CPMI  & 0.765 & 34.6 & 58.8 & 48.2 & 0.14 \textcolor{red}{($\downarrow$86\%)} & 0.01 \textcolor{red}{($\downarrow$99\%)} & \textbf{7.46} & \textbf{6.99} & \textbf{8.75} & \textbf{7.55} \\
CPMI\_Merge1 & 0.766 & 36.3 & \textbf{60.2} & \textbf{49.2} & 0.28 \textcolor{red}{($\downarrow$72\%)} & 0.17 \textcolor{red}{($\downarrow$83\%)} & 3.63 & 3.56 & 4.35 & 3.74 \\
CPMI\_Merge2 & 0.750 & 37.5 & 56.8 & 49.6 & 0.42 \textcolor{red}{($\downarrow$58\%)} & 0.34 \textcolor{red}{($\downarrow$66\%)} & 2.34 & 2.43 & 2.71 & 2.49 \\
CPMI\_Merge3 & \textbf{0.797} & \textbf{39.9} & 56.1 & 48.1 & 0.55 \textcolor{red}{($\downarrow$45\%)} & 0.48 \textcolor{red}{($\downarrow$52\%)} & 1.91 & 1.98 & 2.06 & 1.86 \\
\bottomrule
\end{tabular}%
}
\caption{Efficiency comparison of CPMI and PAV. Computation costs (Time, Generated Tokens) are accumulated over subset of 10K samples.}
\label{app-tab:pav_merg}
\vspace{-0.1in}
\end{table*}

\end{document}